\documentclass[11pt,a4paper]{article}
\usepackage{times,latexsym}
\usepackage{url}
\usepackage[T1]{fontenc}
\usepackage{dirtytalk}
\usepackage{graphicx}
\usepackage{subcaption}  
\usepackage{enumitem}
\usepackage{amsmath}
\usepackage{multirow}
\usepackage{booktabs}
\usepackage{caption}
\usepackage{subcaption}
\usepackage{adjustbox}
\usepackage{wasysym}
\usepackage[acceptedWithA]{tacl2021v1}
\usepackage{ntheorem}
\usepackage{xspace,mfirstuc,tabulary}

\newif\iftaclinstructions
\taclinstructionsfalse 
\iftaclinstructions
\renewcommand{\confidential}{}
\renewcommand{\anonsubtext}{(No author info supplied here, for consistency with
TACL-submission anonymization requirements)}
\newcommand{\instr}
\fi

\iftaclpubformat 

\else

\fi

\title{Preferences for Idiomatic Language are Acquired Slowly \\ --- and Forgotten Quickly: A Case Study on Swedish}

\author{
  Jenny Kunz
  \\
  Department of Computer and Information Science
  \\
  Linköping University
  \\
  \texttt{jenny.kunz@liu.se}
}

\date{}

\begin{document}
\maketitle
\begin{abstract}

In this study, we investigate how language models develop preferences for \textit{idiomatic} as compared to \textit{linguistically acceptable} Swedish, both during pretraining and when adapting a model from English to Swedish. 
To do so, we train models on Swedish from scratch and by fine-tuning English-pretrained models, probing their preferences at various checkpoints using minimal pairs that differ in linguistic acceptability or idiomaticity. For linguistic acceptability, we adapt existing benchmarks into a minimal-pair format. To assess idiomaticity, we introduce two novel datasets: one contrasting conventionalized idioms with plausible variants, and another contrasting idiomatic Swedish with Translationese. 
Our findings suggest that idiomatic competence emerges more slowly than other linguistic abilities, including grammatical and lexical correctness. While longer training yields diminishing returns for most tasks, idiom-related performance continues to improve, particularly in the largest model tested (8B). However, instruction tuning on data machine-translated from English --- the common approach for languages with little or no native instruction data --- causes models to rapidly lose their preference for idiomatic language.
\end{abstract}

\section{Introduction}

Even grammatically correct text can vary in how idiomatic or natural it sounds. Native speakers often prefer certain words or constructions over others, even when both are linguistically acceptable. These preferences influence how fluent or native-like a text \textit{feels}. 

In translation studies, the tendency of translated text to retain source-language features rather than adopting more natural target-language expressions is known as \textit{Translationese} \citep{gellerstam86}. While initially observed in human translations, Translationese is now primarily discussed in relation to machine-translated text. 
We hypothesize that a similar phenomenon arises in language model outputs for non-English languages. Since English dominates LLM pretraining data, transfer effects --- which are generally desirable! --- can introduce anglicized patterns. Furthermore, parts of the training data in lower-resource languages may themselves be translations from English. This influence continues post-training, where steps such as instruction tuning are often performed using machine-translated datasets, as native data is rarely available \citep{muennighoff-etal-2023-crosslingual, holmstrom-doostmohammadi-2023-making, chen-etal-2024-monolingual}.

The idiomaticity of LLM outputs in non-English languages remains however understudied. This property is less straightforward to define or evaluate than grammaticality or factuality, as idiomatic preferences are more subjective and variable. Factors such as age, region, and exposure to English shape what speakers perceive as natural. 

While previous work has examined Translationese in LLM-generated \textit{translations} \citep{raunak-etal-2023-gpts, li2025lostliteralismsupervisedtraining}, to our knowledge, no existing research has investigated general idiomaticity preferences in LLMs. 
Consequently, there are no existing datasets for Swedish that address this issue explicitly. 
To address this gap, we introduce two Swedish datasets designed to probe LLMs’ idiomatic preferences: one contrasts conventionalized idioms with plausible alternatives; the other contrasts Translationese Swedish with idiomatic alternatives. 

We use these datasets to track small (135M, 8B) LLMs' preferences for idiomatic language during pretraining and continued pretraining. Our findings show that idiomatic language is acquired more slowly than other linguistic properties --- slower than syntax, but also slower than appropriate lexical choices. Longer continued pretraining results in steady gains in idiomaticity for English-initialized models, whereas a model trained from scratch plateaus earlier and lower. This suggests that English pretraining not only boosts overall performance but also facilitates the gradual acquisition of idiomatic language use. 

Finally, we examine the effect of instruction-tuning these models with machine-translated data. While performance on the linguistic acceptability benchmarks declines slightly, idiomatic preference drops strongly and early in tuning. This highlights that it is challenging to preserve idiomaticity when tuning the model on translated data. 

\subsection*{Summary}
\noindent We tackle the following two research questions: 

\begin{enumerate}[label=RQ\arabic*, itemsep=0pt, topsep=2pt]
    \item \label{rq:pretraining} How well and how quickly do language models acquire idiomatic preferences, as compared to general linguistic acceptability?
    \item \label{rq:ift} How does instruction tuning with machine-translated data influence a model’s idiomatic preferences? 
\end{enumerate}

\noindent We make the following contributions: 

\begin{enumerate}[leftmargin=*, itemsep=0pt, topsep=2pt]
    \item A manually curated dataset of conventionalized idioms, where one keyword in each expression is replaced with either a semantically similar, high-frequency alternative or a variant inspired by idioms in higher-resourced languages.
    \item A dataset of sentence pairs contrasting Translationese Swedish with idiomatic alternatives.\footnote{Due to copyright restrictions, we cannot release this dataset publicly but it can be easily reproduced from the source book, and we can provide it upon request.}
    \item Empirical findings demonstrating the slow but steady acquisition of preferences for idiomatic language, and their vulnerability to fine-tuning with translated data. 
\end{enumerate}

\section{Related Work}

This work is related to four main strands of research: how linguistic skills are acquired during pre-training (§\ref{sec:skill_acq}), how language transfer and adaptation are done and what the challenges are (§\ref{sec:lang_transfer}), how minimal pair setups are used to probe linguistic competence (§\ref{sec:min_pairs}), and challenges and evaluations of idiomatic language in NLP (§\ref{sec:idiom}).

\subsection{Skill Acquisition during Pretraining}
\label{sec:skill_acq}

Tracking how language models acquire linguistic skills during pretraining has been an active area of research, particularly for encoder-based models, with similar patterns now observed for decoder-based LLMs \citep{wal2025polypythias}.

Prior work often contrasts syntactic skills with higher-level capabilities such as semantic understanding, factual recall, and commonsense reasoning. For instance, \citet{liu-etal-2021-probing-across} find that linguistic knowledge is acquired early and robustly, whereas factual and reasoning abilities emerge later and less reliably across domains. Similarly, \citet{saphra-lopez-2019-understanding} show that syntax is learned earlier than topic-level representations. They also link this to model architecture: lower layers, which converge earlier in training, encode lower-level linguistic information.
\citet{choshen-etal-2022-grammar} find that syntactic abilities are learned in a consistent order across different model sizes and architectures, and that this order even correlates with difficulty for human learners. In a multilingual context, \citet{blevins-etal-2022-analyzing} show that monolingual capabilities are learned early, while cross-lingual generalization develops later.
While much of this work focuses on syntax, semantics, and reasoning, we are, to our knowledge, the first to compare the acquisition of idiomatic versus linguistically generally acceptable language. 

In terms of methodology, many of these studies use probing classifiers \citep{ijcai2018p796} --- small supervised classifiers trained to predict linguistic properties from internal representations --- as in \citet{wal2025polypythias}, \citet{liu-etal-2021-probing-across}, and \citet{blevins-etal-2022-analyzing}. \citet{saphra-lopez-2019-understanding} instead measure representational similarity to task-specific models. Our work is closest in method to \citet{choshen-etal-2022-grammar}, who use a minimal pair setup; an approach we extend to the domain of idiomaticity.

\subsection{Language Transfer}
\label{sec:lang_transfer}

Enabling languages to transfer (linguistic and other) capabilities between each other is a central desideratum of multilingual language models \citep{Costajuss2024ScalingNM, pmlr-v119-hu20b}. Multilingual models are however often evaluated on translated benchmarks, which may not fully capture the quality of their language-specific capabilities,e.g.\ due to missing cultural context \citep{kuulmets-fishel-2023-translated}. Moreover, performance consistency across languages is a known challenge, with models typically performing better on high-resource languages and worse on morphologically rich \citep{arnett-bergen-2025-language} or typologically distant ones \citep{pmlr-v119-hu20b}.

Beyond multilingual models, a growing line of work focuses on post-hoc adaptation, that is, continuing the pretraining of large English-based models on non-English data. For example, the Norwegian Mistral \citep{samuel-etal-2025-small} and AI Sweden LLaMA\footnote{\url{https://my.ai.se/resurser/3070}; Last accessed: Feb 03, 2026} are adapted versions of English-centric models pretrained on data in the languages of Norway and on the mainland Scandinavian languages, respectively. \citet{samuel-etal-2025-small}'s results show that continued pretraining enables the effective use of larger models than what traditional scaling laws would predict, given the small size of available training data. This suggests that leveraging existing cross-lingual representations from English can be highly data-efficient.


\subsection{Minimal Pair Probes}
\label{sec:min_pairs}

Minimal pair probes \citep{linzen-etal-2016-assessing, marvin-linzen-2018-targeted} evaluate linguistic competence by presenting models with pairs of sentences that differ in only a minimal aspect --- typically, one is grammatical or more natural, and the other is not. A model is expected to assign higher probability (i.e., lower perplexity) to the preferred variant \citep{marvin-linzen-2018-targeted}. 
For example, consider this subject–verb agreement pair from \citet{linzen-etal-2016-assessing}: \say{The key is on the table.} vs.\ \say{The key are on the table.}. The first sentence is syntactically correct and should thus be assigned higher probability than the second, incorrect one.

Early work applied this method to recurrent neural networks (RNNs). \citet{linzen-etal-2016-assessing} demonstrated that LSTMs \citep{lstm} can model subject–verb agreement by feeding sentences incrementally and comparing the probabilities of competing verb forms at the critical position. To test syntactic generalization rather than surface memorization, \citet{gulordava-etal-2018-colorless} introduced \say{colorless green ideas}–style stimuli, replacing content words in natural sentences with random words of the same syntactic category. RNNs still performed remarkably well under this setup. \citet{marvin-linzen-2018-targeted} extended this approach to a broader set of syntactic phenomena, including reflexive binding and negative polarity licensing.

With the rise of Transformer-based models \citep{NIPS2017_3f5ee243} like BERT \citep{devlin-etal-2019-bert}, minimal pair probing has been adapted to masked language modeling. \citet{goldberg2019assessingbertssyntacticabilities} test BERT’s syntactic knowledge using cloze-style minimal pairs, while \citet{talmor-etal-2020-olmpics} apply similar setups to probe commonsense reasoning, including age and object comparisons.

Several benchmark datasets have also been created to systematize this methodology. BLiMP (Benchmark of Linguistic Minimal Pairs) \citep{warstadt-etal-2020-blimp-benchmark} provides automatically generated minimal pairs targeting twelve linguistic phenomena, such as determiner–noun agreement, anaphora binding, and ellipsis. These benchmarks provide a controlled setting for evaluating specific linguistic capabilities and are especially suited to tracking how models acquire such skills over training.

\subsection{Idioms}
\label{sec:idiom}

Conventionalized idioms are fixed expressions whose meanings deviate from their literal forms. As culturally grounded and figurative constructions, idioms can be challenging for both human learners and language models. Figurative language --- including idioms --- has long been recognized as a bottleneck in natural language understanding \citep{shutova2011computational}. 

While some idioms have cross-lingual equivalents due to shared religious or literary heritage (e.g., the biblical idiom \say{cast pearls before swine} has the Swedish equivalent \say{kasta pärlor åt svin}), others are culture-specific. Literal translations frequently produce unnatural output, even when the intended meaning is clear. Swedish idioms such as \say{gå åt skogen} (“go to the forest,” meaning “go wrong”) or \say{gå av stapeln} (\say{to be launched}, originally a maritime term) reflect ties to nature, local customs, and daily life.

This cultural specificity makes idioms particularly challenging for multilingual LLMs. In a qualitative study, \citet{pedersen-etal-2025-evaluating} show that LLMs explain English metaphors more effectively than Danish-specific ones, raising concerns about linguistic diversity in LLMs. Idioms are also a well-known challenge in machine translation, where literal renderings are common \citep{fadaee-etal-2018-examining, dankers-etal-2022-transformer, liu-etal-2023-crossing}, and a lack of correct interpretation even affects downstream tasks tasks such as textual entailment \citep{chakrabarty-etal-2021-figurative, stowe-etal-2022-impli}.

Idioms are more difficult that other linguistic phenomena even in human second-language acquisition. Learners tend to avoid using idioms, especially opaque or culture-bound ones, and prefer transparent idioms or those with equivalents in their native language \citep{laufer2000avoidance, tarabrina2019addressing}. As \citet{pinnavaia2002grammaticalization} notes, idioms are the \say{patrimony of a culture}, deeply embedded in specific sociolinguistic and geographical contexts.

Most NLP work on idioms focuses on detection \citep{tayyar-madabushi-etal-2022-semeval, yayavaram-etal-2024-bert}, disambiguation \citep{kurfali-ostling-2020-disambiguation, garcia-etal-2021-probing}, translation \citep{baziotis-etal-2023-automatic}, or explanation \citep{gluck2025clixcrosslingualexplanationsidiomatic, pedersen-etal-2025-evaluating}. In contrast, we address active idiomatic competence: the ability to produce idiomatic expressions from inputs that give a definition or explanation of the idiom. This represents a higher bar than translating or explaining a given idiom as it ensures that the model is able to actively \textit{produce} the idiom, while a correct interpretation can often be inferred from context. Perhaps most similar to our work is a subset of the Norwegian benchmark NorEval \citep{mikhailov-etal-2025-noreval}, where the last word of an idiom is cut off and has to be predicted. The EuroEval benchmark \citep{nielsen-2023-scandeval} even includes a multiple-choice variant of this dataset, including LLM-generated incorrect alternatives. 

Finally, \citet{liu-lareau-2024-assessing} find that CamemBERT predicts idiomatic tokens (in French) \textit{more} reliably than literal ones, especially in longer expressions. This suggests that idioms, once learned, function as cohesive units with strong internal constraints, making them more predictable despite their semantic opacity.

\section{Minimal Pair Benchmarks}

We develop two sets of minimal pair benchmarks: A baseline with \textit{linguistic acceptability} datasets from existing work (§\ref{sec:la}), and one with our own datasets for \textit{idiomatic preferences} (§\ref{sec:ip}). 

\subsection{Baseline: Linguistic Acceptability}
\label{sec:la}

As a baseline for monitoring the acquisition of idiomatic preferences, we construct a linguistic acceptability benchmark. We follow the minimal pair paradigm, where each item consists of one correct sentence and a closely related ungrammatical counterpart. This formulation allows us to assess whether models prefer the well-formed sentence over its erroneous variant --- an approach grounded in prior work on linguistic minimal pairs \citep{linzen-etal-2016-assessing, marvin-linzen-2018-targeted}. 

In contrast to English, subject-verb agreement (the most common target in minimal pair probing) does not exist in modern Swedish, so that we cannot make use of this classical task. Instead, we focus on naturally occurring and synthetically induced grammatical errors that reflect Swedish morphosyntax and word order phenomena. 

\subsubsection{Naturally Occurring Errors}

The DaLAJ (Dataset for Linguistic Acceptability Judgments) benchmark \citep{volodina-etal-2023-dalaj} provides a basis for constructing minimal pairs from real-world learner language. It is based on essays written by second-language learners of Swedish across different proficiency levels. 
Each sentence in the dataset has been manually corrected, modified so that it contains exactly one error, and annotated for the type of linguistic error. 

As the DaLAJ dataset contains sentences with error annotations, but not a direct reference to the correction, we need to align each incorrect sentence with its correct correspondence to construct minimal pairs. We do so by searching the correct sentence with the smallest Levenshtein distance \citep{levenshtein1966binary} for each incorrect sentence. 
We concatenate the original training, validation and test splits to maximize coverage, resulting in 20,000 sentence pairs (with 6,500 unique correct sentences) of which 2122 are \textit{punctuation} errors, 3924 are  \textit{orthography} errors, 3754 are \textit{lexical choice} errors, 6482 are \textit{morphological} errors and 4666 are \textit{syntax} errors. 

\subsubsection{Synthetic Corruptions}

The ScaLA dataset \citep{nielsen-2023-scandeval} provides another source of acceptability judgments by using synthetic corruption of grammatical sentences. In this dataset, well-formed sentences are modified by introducing controlled perturbations: Either two adjacent tokens are swapped (called \textit{flip neighbours} in this paper), or a single token is deleted (\textit{delete}).
For each corrupted sentence, we identify its closest grammatical source sentence, again using the minimum Levenshtein distance, to form minimal pairs. 
As with DaLAJ, we combine all splits (train, validation, test) to construct the final set, resulting in 552 pairs for the \textit{flip neighbors }and 601 pairs for the \textit{delete} category. 

\subsection{Idiomatic Preferences}
\label{sec:ip}

As no existing benchmarks evaluate a model’s preference for idiomatic language use in Swedish, we introduce two new datasets using the same minimal pair setup as in the syntactic tasks. 

\subsubsection{Swedish Conventionalized Idioms}
\label{secsec:idioms_ds}

We construct a benchmark of Swedish idioms initially based on a completion task, and then reformulated into minimal pairs. Our aim is to test whether models prefer the idiomatic keyword over a plausible but non-idiomatic alternative. We briefly describe the creation process here; more details and examples can be found in Appendix~\ref{app:dataset_construction}. 

\paragraph{Data Collection.}
We manually curate a list of Swedish idioms and their definitions by browsing relevant Wiktionary categories and lists\footnote{\url{https://sv.wikipedia.org/wiki/Lista_över_svenska_idiomatiska_uttryck}, \url{https://sv.wiktionary.org/wiki/Appendix:Idiomatiska_uttryck/Svenska}, \url{https://sv.wiktionary.org/wiki/Kategori:Svenska/Idiomatiskt}; Last access: Feb. 03, 2026} and selecting those that:
\begin{itemize}[leftmargin=*, itemsep=0pt, topsep=2pt]
\item can be reasonably framed as a cloze task with a single missing word;
\item have a clearly identifiable idiomatic answer (without being overly obvious or trivially inferrable from context). 
\end{itemize}
This resulted in 479 idioms. 
We formulated each instance as a short sentence or question that includes the idiom's definition or explanation and is aimed at eliciting a specific idiomatic keyword. An example for such a sentence is: \textit{Någon som är väldigt rik är rik som ett troll} (\say{Someone who is very rich is rich like a troll}). 
In this formulation, clarity was prioritized over syntactic or lexical diversity. 
The dataset was initially compiled by one annotator and then reviewed by two others for the full dataset Idioms$_\text{all}$. 

\paragraph{Minimal Pair Construction.}
For each idiom, we manually construct a plausible but less idiomatic alternative keyword. Whenever possible, the less idiomatic option (the distractor) is based on a similar idiom in a closely related high-resource language, such as English or German. For example, in Swedish the idiom is \textit{jämföra äpplen och päron} (\say{comparing apples and pears}), whereas in English it is \say{comparing apples and oranges.} Thus, \textit{apelsiner} (\say{oranges}) becomes the distractor. 
To find such counterparts, annotators use their linguistic knowledge, web search, and resources such as the Oxford Dictionary of Idioms. When no direct cross-lingual idiom exists, the distractor is chosen from semantically and syntactically plausible alternatives in the same lexical category. Absurd alternatives are discouraged. 

We then use the chatbot assistant Claude \citep{claude3} to rewrite prompts into minimal pair sentences with both the intended idiomatic keyword and the distractor inserted. These keywords can appear in any part of the sentence. 
All generated pairs are manually checked and revised to ensure fluency and idiomatic language. 

\paragraph{Challenge Subset.}
We additionally construct a challenge subset Idioms$_\text{challenge}$ of idioms that: 
\begin{enumerate}[leftmargin=*, itemsep=0pt, topsep=2pt]
\item have no exact counterparts in either English or German;
\item lack strong surface cues that would make them easy to guess. 
\end{enumerate}
To verify the absence of equivalents in English and German, our annotators conduct web searches and consult the Oxford Dictionary of Idioms. This subset is reviewed and enhanced by three annotators, including two native speakers of German. 



\paragraph{Release.}
We make this dataset publicly available on the Hugging Face Hub under the identifier \textit{liu-nlp/swedish-idioms} and on \url{https://github.com/jekunz/idiomatic-language}. 

\subsubsection{Translationese}
\label{subsubsub:translationese}

\textit{Translationese} refers to the systematic influence of a source language on the target language in translation. The concept was introduced by \citet{gellerstam86}, who defined it as a set of linguistic patterns that distinguish translated text from native language use. 

According to \citet{baker}, professional translations often exhibit a strong preference for conventional grammaticality and style. This bias may lead to stylistically flattened or simplified output that, paradoxically, results in lower perplexity compared to more idiomatic or natural native expressions, making a dataset comparing Translationese with non-Translationese more challenging. 

\paragraph{Dataset Construction.}
We base our dataset on \citet{katourgi2022svenskan}, a popular science book that systematically contrasts \say{naturally occurring} Swedish translations influenced by English with more natural Swedish formulations proposed by the author. 
For example, it contrasts the Translationese sentence \textit{Låt oss ta en titt}, which is a literal translation of the English sentence \say{Let us take a look}, with the more idiomatic \textit{Kom så kollar vi} (Literally: \say{Come then we check}).
We transcribe all sentence pairs from the book into a structured dataset. In cases where two alternative idiomatic versions are provided, we include both, resulting in a total of 967 sentence pairs. We denote this dataset Translationese$_\text{all}$. 

\paragraph{Subset for More Reliable Evaluation.} 
It is important to note that the examples from the book are not per se minimal: the idiomatic rewritings tend to be shorter and more freely adapted than the original Translationese versions. On average: 
\begin{itemize}[leftmargin=*, itemsep=0pt, topsep=2pt]
\item The Translationese variants contain 5.09 whitespace-tokenized tokens.
\item The idiomatic alternatives contain 4.41 tokens.
\end{itemize}
Even though we normalize the perplexity by sequence length (see Section~\ref{secsec:metrics}), this difference can add systematic biases, e.g.\ by longer sequences containing more easy-to-predict function words. In subword tokenizers (which most current models, including those on our experiments, use), later subwords in a word have lower perplexities because they have a strong cue, that is, the preceding subword(s) \citep{chang-etal-2024-characterizing}. 

For a fair minimal pair evaluation, we therefore extract a filtered subset Translationese$_\text{filtered}$ based on the following two criteria:
\begin{enumerate}[leftmargin=*]\setlength\itemsep{0pt}
\item The two variants have the same length according to the tokenizer of the respective model. 
\item The idiomatic variant is clearly preferable based on general Swedish language norms, even without additional context. This criterion has been implemented with one primary rater and one reviewer of the annotations. 
\end{enumerate}
The intersection of both subsets results in 98 samples for the SmolLM models and 134 samples for the Llama model.\footnote{Note that due to the different subsets resulting from different tokenizers, the results of the SmolLM models are not directly comparable to the Llama model. As we focus on intra-model comparisons, this is of less importance in this paper. For inter-model comparisons, basing the length restriction on whitespace tokenizer can be a compromise. This approach results in a larger dataset but retains a stronger bias towards the Translationese samples than the model-tokenizer based filtering, as visible in Table~\ref{tab:translationese_ablations} in Appendix~\ref{app:tokenizer}.} 

\paragraph{Data Availability.}
Because the dataset is derived from a copyrighted book, we are unable to publish it freely. However, it can be reproduced by transcribing all sentences from \citet{katourgi2022svenskan}; we share the indices of the samples included in the challenge set at \url{https://github.com/jekunz/idiomatic-language}. We are able to share both the full and filtered datasets upon request with researchers with legal access to the source material. 

\subsection{Annotators}

The primary annotators and raters of the datasets are two student assistants (paid after collective agreement) who are native-level speakers of Swedish and have undergone basic training in linguistics in their Cognitive Science study program. The first author reviewed all annotations. 

\section{Experimental Setup}

All models and checkpoints used in our experiments are available on the HuggingFace Hub in the collection \url{https://huggingface.co/collections/jekunz/idiomatic-language-acquisition}. The code is available at \url{https://github.com/jekunz/idiomatic-language}. 

\subsection{Training} 
To answer our research questions, we train in two phases: (continued) pre-training (\ref{rq:pretraining}) and instruction tuning (\ref{rq:ift}). 

\paragraph{Pretraining (\ref{rq:pretraining}). }
We use the Swedish portion of the deduplicated FineWeb2 dataset \citep{penedo2025fineweb2pipelinescale}, which contains approximately 25 billion tokens, to train two models with 135M parameters based on the SmolLM2 architecture \citep{allal2025smollm2smolgoesbig}: \textit{SmolLM$_\text{CPT}$} is a continued pretraining (CPT) of the original SmolLM2 135M model, which was trained on the English Fineweb dataset \citep{penedo2024the}.  \textit{SmolLM$_\text{scratch}$} is trained only on Swedish from random initialization. 

Both models are trained with identical hyperparameters: we use the AdamW optimizer \citep{loshchilov2018decoupled}, a cosine learning rate scheduler with a warmup rate of 5\%, an effective batch size of 256, and a context window size of 8192 tokens (the same as in the SmolLM2 pre-training). 
Checkpoints are saved every 200 training steps, corresponding to every 200 million tokens processed. For training on the full 25 billion tokens, the models required approximately 720 GPU hours on NVIDIA A100 40GB cards. 

\paragraph{Instruction Tuning (\ref{rq:ift}). } To see the effect of instruction tuning with translated data, we translate Smol-Smoltalk \citep{allal2025smollm2smolgoesbig}, a synthetic instruction-tuning dataset specifically adapted for models with less than 1B parameters, to Swedish using Gemma-27B \citep{gemma}\footnote{We choose Gemma due to its multilingual capabilities while being relatively cheap to use: The translation of the dataset cost 144 A100 hours.}. In the prompt, we instruct the model to phrase the translations as naturally and idiomatically in Swedish as possible. A manual inspection of 20 translated conversations indicated reasonable translation quality for current standards, with little obvious Translationese. 
We train the SmolLM models on the dataset (456,780 conversations) for one epoch with the same parameters as for the pre-training, requiring 12 A100 hours per model. 

\subsection{Existing Models}

We evaluate AI-Sweden-Llama, a version of Llama 3 8B \citep{llama3} further trained on the Swedish, Danish, and Norwegian subsets of the Nordic Pile \citep{ohman2023nordicpile12tbnordic}. The model has ten publicly released checkpoints recorded after 1500, 2700, 3900, 5325, 6550, 8200, 11525, 14375, 16000, and 18833 training steps, allowing us to tackle \ref{rq:pretraining}. 
For comparison, we also evaluate the base Llama 3 8B model, which is officially an English model, as checkpoint 0. A certain exposure to Swedish already in the base model's pre-training is however likely. The model was instruction-tuned on a Swedish translation of OpenHermes 2.5 \citep{OpenHermes25}, produced using GPT-3.5\footnote{Information via personal communication with model creator Tim Isbister.}, which allows us to tackle \ref{rq:ift}.

\subsection{Metric}
\label{secsec:metrics}

To evaluate model preference, we compute the average per-token perplexity of the full sentence $x_{1:N}$ of length $N$: 
\vspace{-0.3em}
\begin{equation*}
\text{ppl}(x) = \exp\!\left(-\frac{1}{N} \sum_{t=1}^{N} \log p(x_t \mid x_{<t}) \right).
\end{equation*} 
\vspace{-0.5em}

For each minimal pair, we calculate perplexity for both sentences.
The sentence with the lower perplexity is taken to be the model’s preferred choice.
Accuracy is then defined as the proportion of $\hat{y}$ in which the model prefers the better (i.e., grammatical or more idiomatic) sentence $x_{\text{better}}$:
\vspace{-0.3em}
\begin{equation*}
\hat{y} =
\begin{cases}
1 & \text{if } \text{ppl}(x_{\text{better}}) < \text{ppl}(x_{\text{worse}}) \\
0 & \text{otherwise.}
\end{cases}
\end{equation*}

\section{Results and Analysis}
To answer \ref{rq:pretraining}, we analyze how models acquire idiomatic language during pre-training in Section~\ref{sec:acquisition}. 
Then, tackling \ref{rq:ift}, we examine the impact of instruction fine-tuning on these capabilities in Section~\ref{sec:ift}. Finally, we analyze \textit{why} idiomatic language preferences are more sensitive than general linguistic competence in Section~\ref{sec:why}. 

\begin{table*}[ht]
    \centering
    \adjustbox{max width=0.75\textwidth}{
    \begin{tabular}{llcccccc}\toprule
    & & \multicolumn{2}{c}{SmolLM$_\text{scratch}$} & \multicolumn{2}{c}{SmolLM$_\text{CPT}$} & \multicolumn{2}{c}{AI Sweden Llama}
    \\\cmidrule(lr){3-4}\cmidrule(lr){5-6}\cmidrule(lr){7-8}
           && Before & After & Before & After & Before & After\\\midrule
    \multirow{6}{*}{\rotatebox{90}{Final Accuracy}}
    & Idioms$_\text{all}$ & 63.17 & 48.53 & 81.79 & 68.41 & 94.14 & 88.91 \\
    & Idioms$_\text{challenge}$ & 62.00 & 40.00 & 77.00 & 59.00 & 92.00 & 84.00 \\\cmidrule(lr){3-8}
    & Translationese$_\text{all}$ & 46.94 & 36.91 & 47.36 & 34.33 & 48.91 & 35.36 \\
    & Translationese$_\text{filtered}$ & 58.20 & 50.00 & 69.40 & 50.00 & 82.08 & 59.70 \\\cmidrule(lr){3-8}
    & DaLAJ$_\text{lexical}$ & 77.19 & 72.58 &  85.58 & 80.58 &  90.17& 87.45 \\
    & DaLAJ$_{\text{others}}$   & 88.82 & 86.60 &  92.65 & 92.35 &  94.69 & 94.10 \\\cmidrule(lr){3-8}
    & ScaLA (avg.)           & 90.03 & 86.38 & 95.82 & 93.92 &   98.61& 98.19  \\\bottomrule
    \end{tabular}
    }
    \caption{Final accuracy after (continued pre-) training on Swedish. We report results \textit{before} and \textit{after} instruction tuning for all models and tasks.}
    \label{tab:final-accuracy}
\end{table*}

\begin{table}[ht]
    \centering
    \adjustbox{max width=0.49\textwidth}{
    \begin{tabular}{lccc}\toprule
    & SmolLM$_\text{scratch}$ & SmolLM$_\text{CPT}$ & AI Swe.\ Llama \\\midrule
    Idioms$_\text{all}$           & 6200 & 11600 & 3900 \\
    Idioms$_\text{chall.}$     & 8000 & 11400 & 3900 \\\cmidrule(lr){2-4}
    Transl.$_\text{all}$   & 0    & 3000  & 3900 \\
    Transl.$_\text{filtered}$& 0    & 1200  & 5325 \\\cmidrule(lr){2-4}
    DaLAJ$_\text{lexical}$        & 4400 & 2400  & 1500 \\
    DaLAJ$_{\text{others}}$*      & 3933 & 1266  & 0 \\\cmidrule(lr){2-4}
    ScaLA (avg.)                  & 3700 & 1300  & 0 \\\bottomrule
    \end{tabular}
    }
    \caption{Training step at which each model reaches 95\% of its total accuracy. (* We exclude DaLAJ punctuation here because it is 0 for all models.)}
    \label{tab:95step}
\end{table}

\begin{figure*}[htbp]
    \centering
    \begin{subfigure}[b]{0.48\textwidth}
        \includegraphics[width=\textwidth]{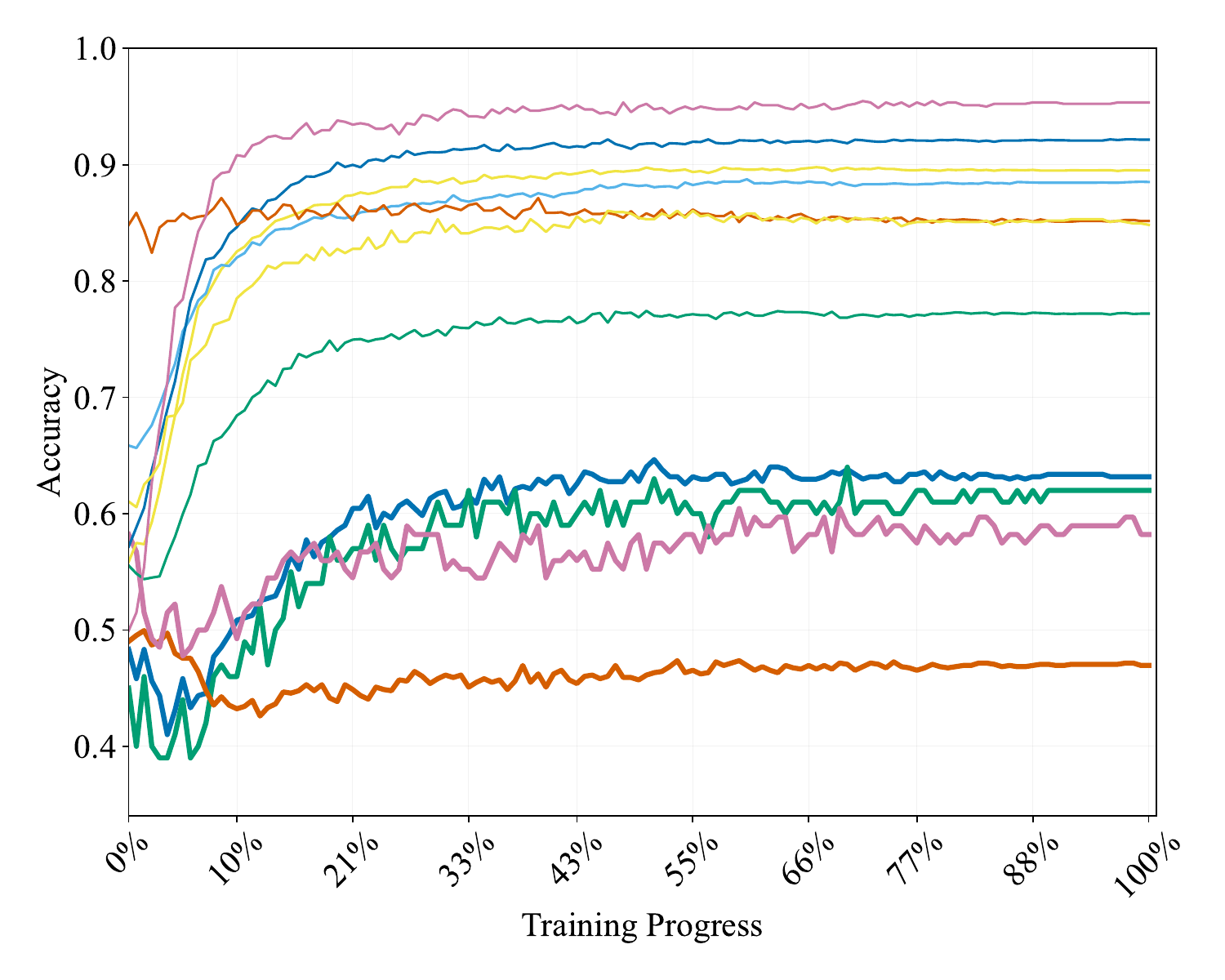}
        \caption{SmolLM2: Training from scratch.}
        \label{fig:scratch}
    \end{subfigure}
    \hfill
    \begin{subfigure}[b]{0.48\textwidth}
        \includegraphics[width=\textwidth]{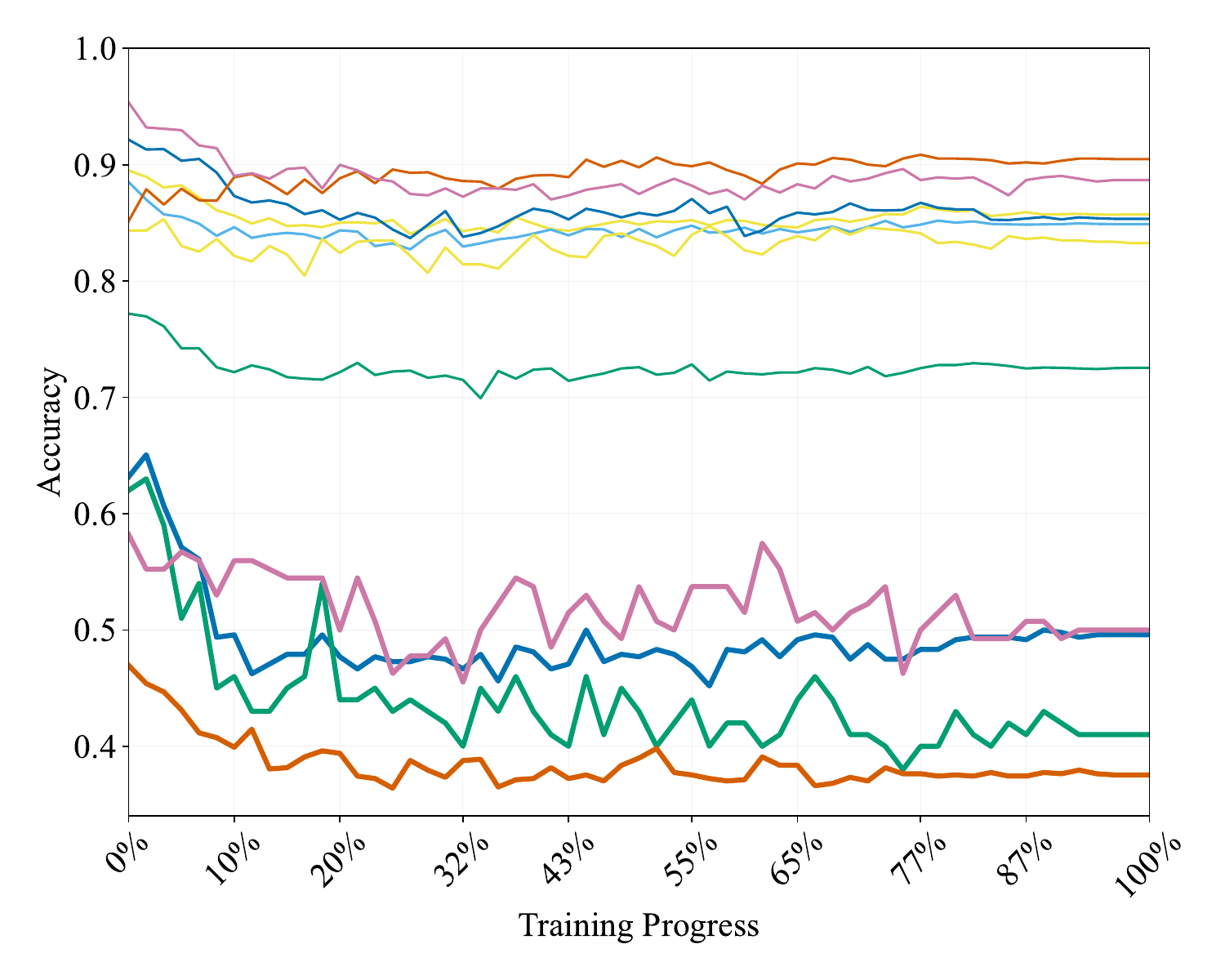}
        \caption{SmolLM2 from scratch: Instruction tuning.}
        \label{fig:scratch-ift}
    \end{subfigure}

    \begin{subfigure}[b]{0.48\textwidth}
        \includegraphics[width=\textwidth]{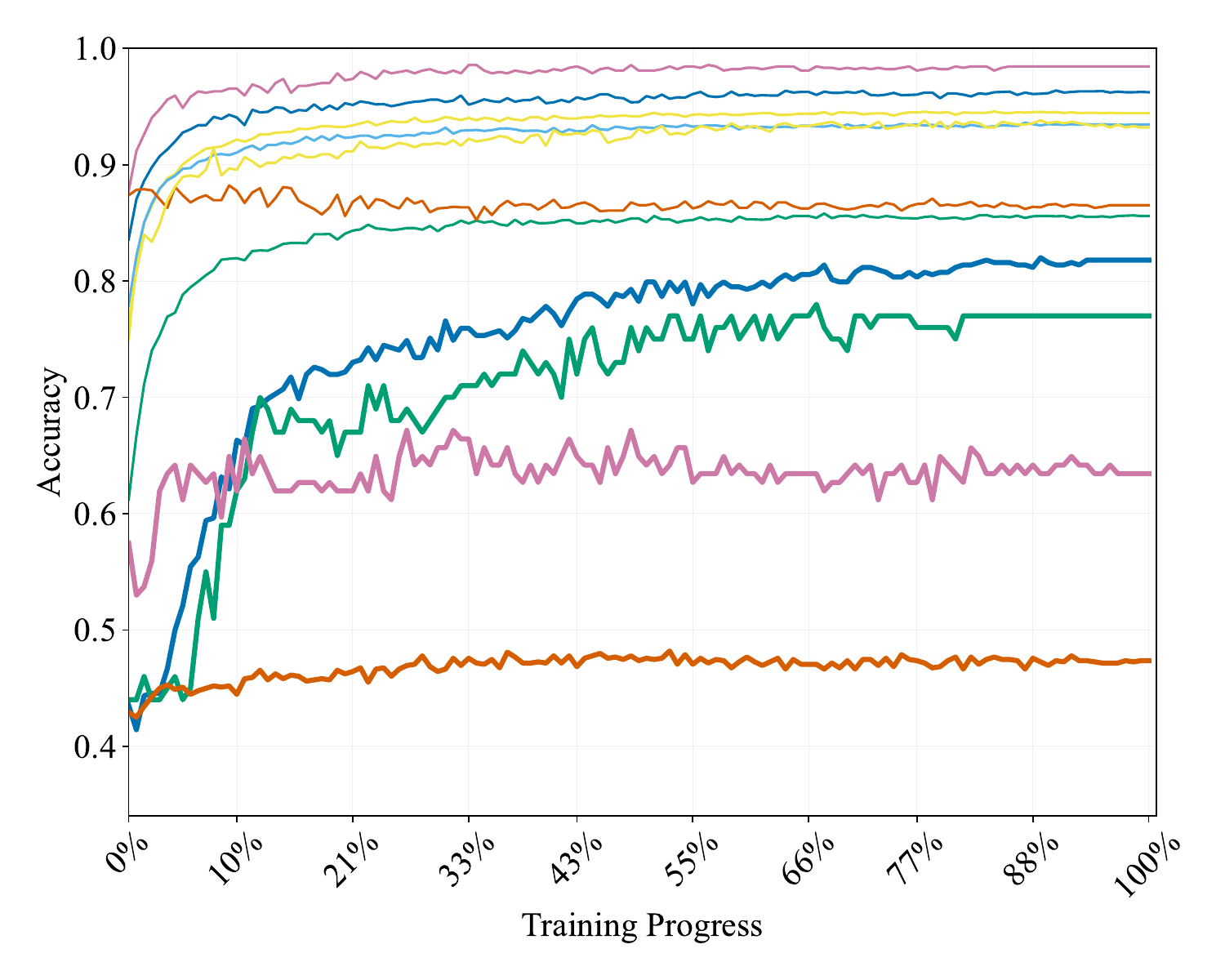}
        \caption{SmolLM2: Continued pretraining from English.}
        \label{fig:cpt}
    \end{subfigure}
    \hfill
    \begin{subfigure}[b]{0.48\textwidth}
        \includegraphics[width=\textwidth]{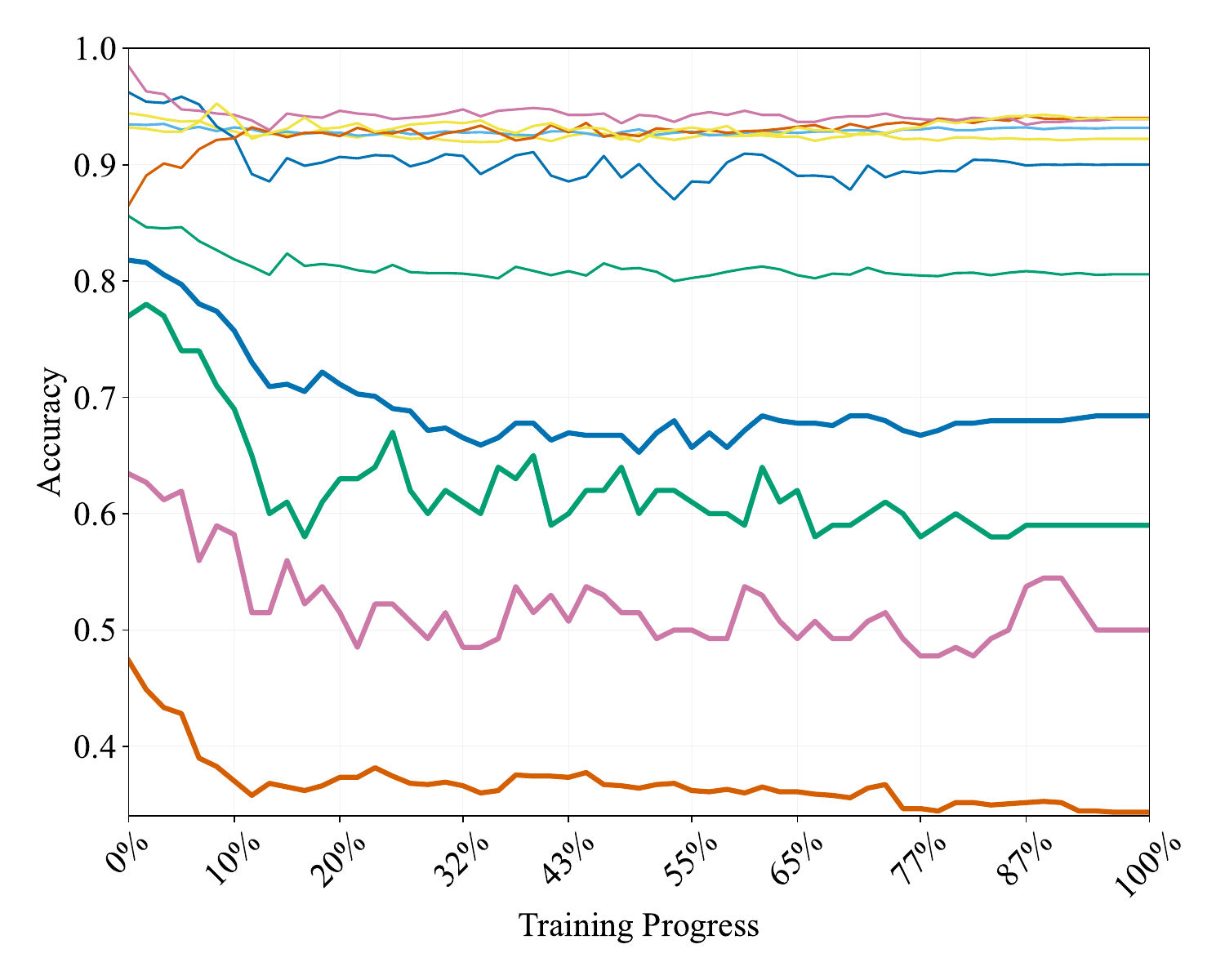}
        \caption{SmolLM2 CPT from English: Instruction tuning.}
        \label{fig:cpt-ift}
    \end{subfigure}

    \begin{subfigure}[b]{0.48\textwidth}
        \includegraphics[width=\textwidth]{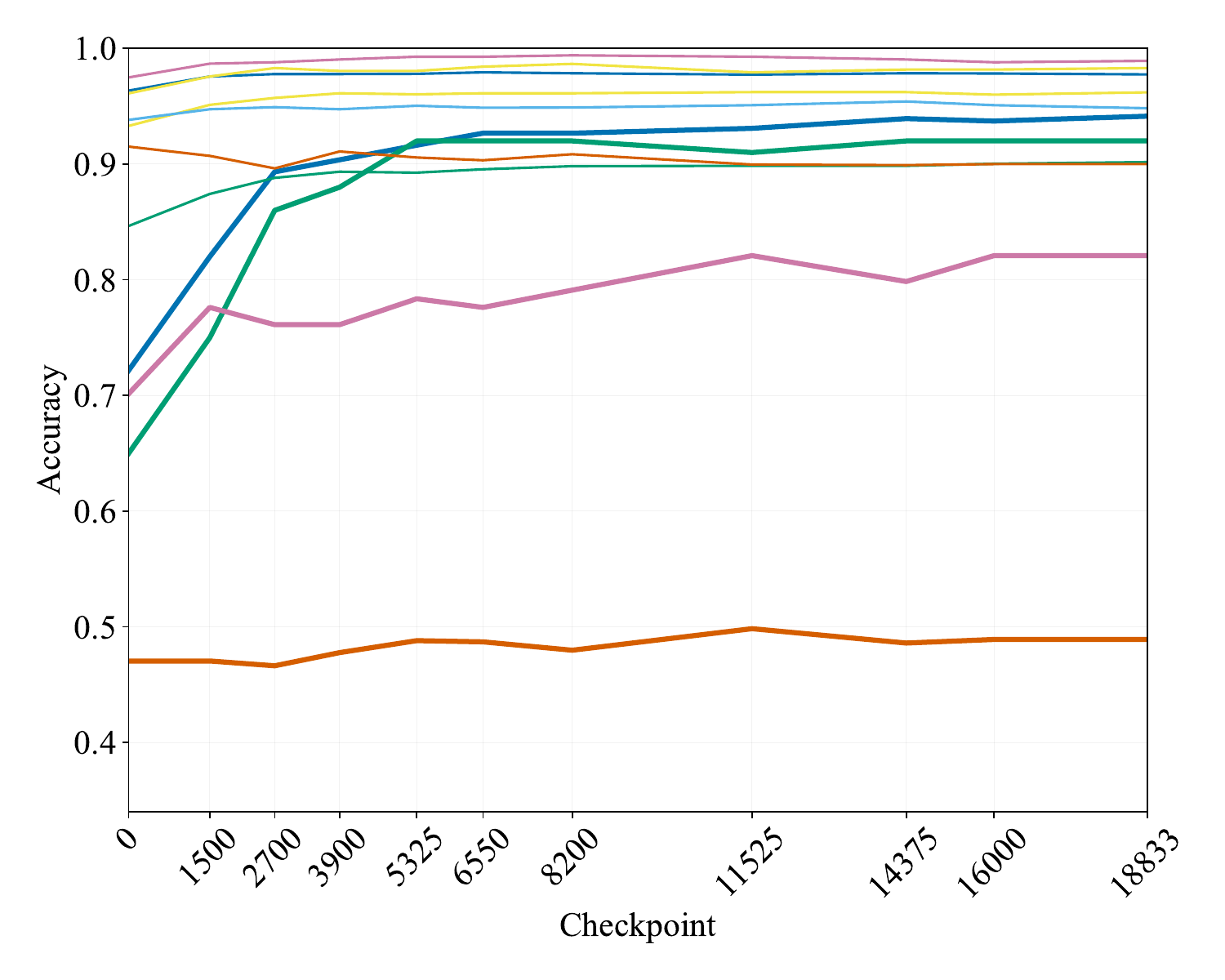}
        \caption{AI Sweden LLaMA 8B: Continued pretraining.}
        \label{fig:llama}
    \end{subfigure}
    \hfill
    \begin{subfigure}[b]{0.45\textwidth}
        \includegraphics[width=\textwidth]{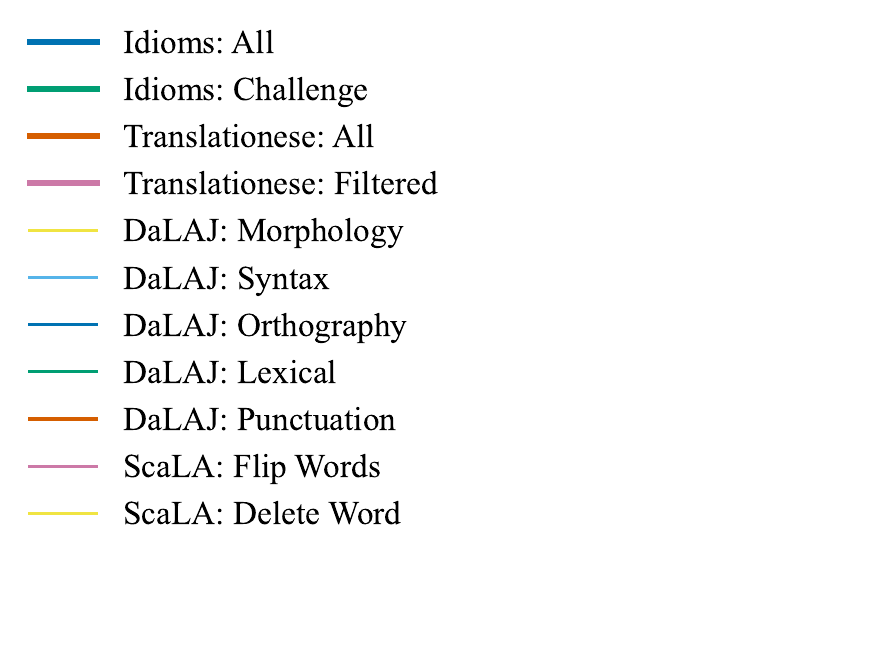}
        \caption{Legend for all Figures: Datasets and subsets.}
        \label{fig:legend}
    \end{subfigure}

    \caption{Accuracy across training checkpoints for all SmolLM2 135M variants and for AI Sweden LLaMA 8B. The subfigures on the left are for (continued) pre-training with plain text, the subfigures on the right for instruction tuning. Subfigure~\ref{fig:legend} contains the legend for all plots. }
    \label{fig:all-models-combined}
\end{figure*}

\subsection{\textit{Acquisition} of Linguistic Skills (\ref{rq:pretraining})}
\label{sec:acquisition}

All three models are better at distinguishing linguistically (syntactically, orthographically and lexically) correct from incorrect Swedish than they are at distinguishing idiomatic from unidiomatic language: The scores for the linguistic acceptability datasets DaLaJ and ScaLA in Table~\ref{tab:final-accuracy} are higher across models than for the Idioms and Translationese datasets. 
Idiomaticity remains challenging to acquire and improves slowly during language adaptation, as visible in the left column in Figure~\ref{fig:all-models-combined} (Subfigures~\ref{fig:scratch}, \ref{fig:cpt} and \ref{fig:llama}). 

We observe consistent performance gains with English pretraining compared to SmolLM$_\text{scratch}$. English pretraining does not hinder the learning of idiomaticity (or any other linguistic skill included in our evaluations) but provides a better base for learning fine-grained Swedish capabilities. 

The base LLaMA 3 model (checkpoint 0 in Figure~\ref{fig:llama}), despite being trained primarily on English, already performs well across linguistic acceptability tasks: It has more than 95\% of its final accuracy on the DaLaJ and ScaLA benchmarks already before adaptation, as seen in Table~\ref{tab:95step}. However, its lexical and idiomatic competence is markedly weaker, and requires continued pretraining to acquire, as we can see in Figure~\ref{fig:llama}. 

\subsubsection{Idioms}
\label{secsec:idioms}

This section examines how models acquire the conventionalized idioms of the \textit{Idioms} datasets, and how this differs from their ability to learn lexical or syntactic norms. 

\paragraph{Idioms versus Lexical Correctness.} Interestingly, the LLaMa model not only catches up in idiom competence but \textit{surpasses} its performance on the lexical acceptability benchmark (DaLAJ$_\text{lexical}$) relatively early in training (by step 2,700). These findings are in line with \citet{liu-lareau-2024-assessing}, who show that idiomatic expressions in French are easier to predict in context, due to their strong internal constraints. However, this pattern does not hold for the smaller models, and suggests that both a high learning capacity (parameter count) and extensive Scandinavian-language training data are key enablers for learning conventionalized idioms. 

\paragraph{Learning Curves.} For AI Sweden LLaMA and SmolLM$_\text{CPT}$, idiom preferences continue to improve steadily throughout training, even as performance on other tasks (including DaLAJ$_\text{lexical}$, the most similar task) plateaus. This suggests that the Idioms dataset requires more training data and longer exposure to perform well on, but remains learnable with sufficient capacity and training. 
In contrast, SmolLM$_\text{scratch}$ plateaus early on the Idioms tasks and reach significantly lower performance levels, suggesting that the rarer idioms are not successfully learned. 

The commonly observed breakpoint around 2–5B tokens, which is linked to the emergence of circuits responsible for core linguistic structure \citep{tigges2024llm, wal2025polypythias}, is reflected in the linguistic acceptability tasks, where performance jumps occur early (roughly at 10–25\% of training). Idiom acquisition, by contrast, lacks a sharp inflection point and instead improves gradually, with substantial gains continuing well beyond the convergence of other tasks. This suggests that idioms require sustained exposure and are acquired more slowly than structural aspects of language. 

\begin{figure}
    \includegraphics[width=0.5\textwidth]{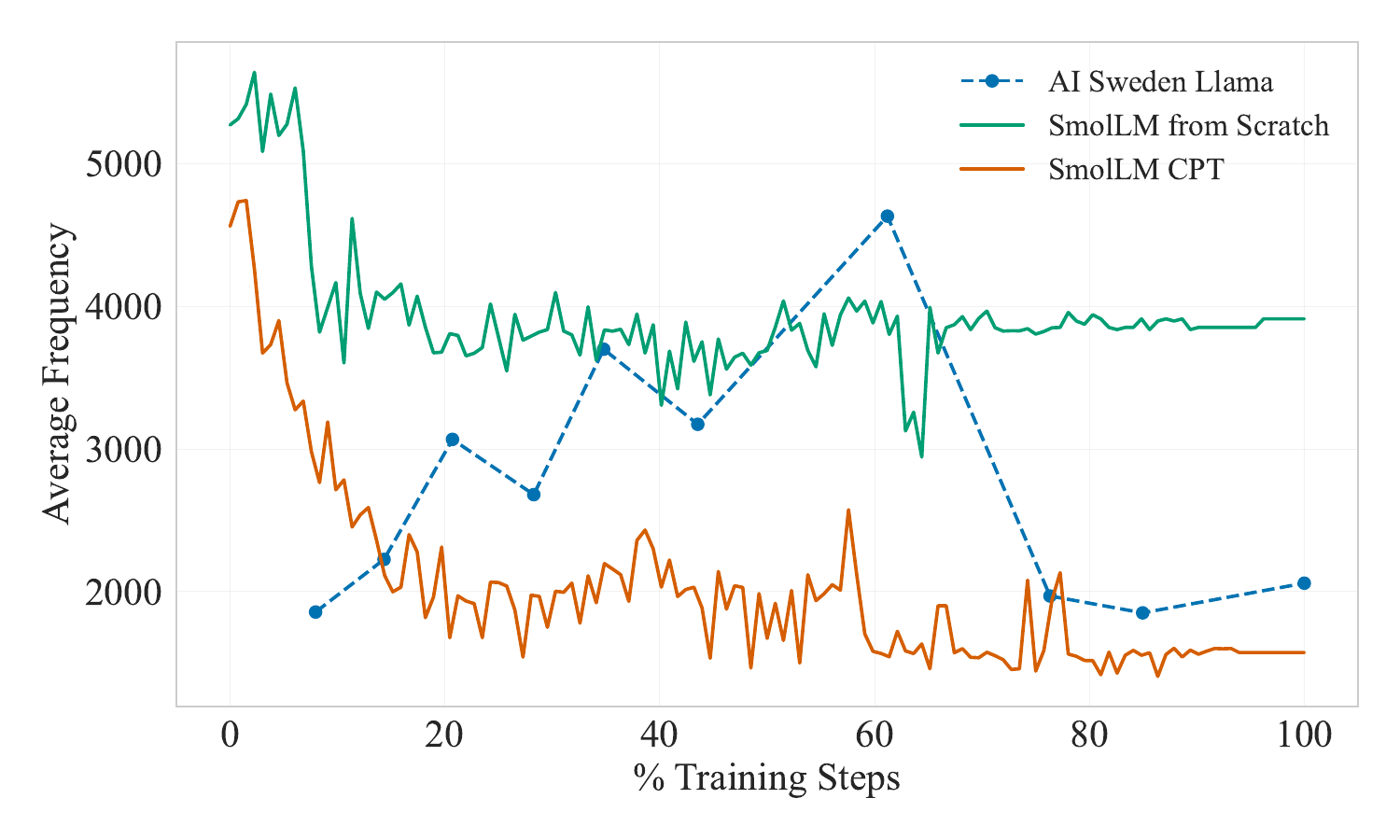}
    \caption{Average frequencies of idioms mispredicted at each checkpoint in the FineWeb2 corpus. Average over all idioms in the dataset: 5766.}
    \label{fig:frequencies}
\end{figure}

\paragraph{Effect of Idiom Frequency.}
We analyze the frequency of idioms in the FineWeb2 corpus using a lexical search, identifying each idiom by the shortest substring that likely uniquely identifies it.\footnote{This method is imperfect: it captures only a subset of instances for compositional idioms and may fail to fully exclude literal uses. A more recall-optimized strategy would however require careful engineering for each idiom and is outside the scope of this article. } For each training checkpoint, we collect the idioms predicted incorrectly and report their average frequency in FineWeb2 in Figure~\ref{fig:frequencies}.

For both SmolLM models, incorrectly predicted idioms initially have a higher average frequency and then gradually converge towards a lower frequency. 
The model trained from scratch converges faster and stabilizes at a higher frequency, whereas the continually pre-trained model’s curve decreases slightly almost until the end. These trends closely mirror the training curves for idioms shown in Figures~\ref{fig:scratch} and \ref{fig:cpt}: as performance improves, the average frequency of mispredicted idioms declines. In contrast, AI Sweden Llama shows a different pattern: the average frequency of mispredicted idioms is low at the earliest checkpoints, increases at intermediate steps, and then declines again toward the final checkpoints. Moreover, idioms that remain incorrectly predicted after full training are not less frequent than those missed at the first checkpoint: the average frequency of mispredicted idioms is 1857 at the first and 2059 at the final checkpoint. 
 This difference could have two explanations. First, AI Sweden Llama is trained on a different, non-public corpus, which may have a distinct idiom distribution. Second, its higher learning capacity and overall better performance mean that fewer idioms are mispredicted, so other factors beyond frequency may play a larger role. 
Examining the mispredicted idioms for AI Sweden Llama reveals a pattern: many rarer, non-compositional idioms that are initially missed are eventually learned, likely because their strong lexical cues make them easier to memorize from limited examples compared to more compositional idioms. For instance, the idiom \textit{inte för allt smör i Småland} (literally, \say{not for all butter in Småland}, roughly equivalent to English \say{not for all the money in the world}) appears only 1,269 times in the corpus, making it a quite rare idiom, yet its highly non-compositional form and distinctive word sequence allow the model to learn it with relatively small exposure: It is mispredicted at the first checkpoint, but then memorized. 

\paragraph{Swedish-specific idioms} (Idioms$_\text{challenge}$) show a learning trajectory similar to that of Idioms$_\text{all}$, although with slightly lower overall performance. This suggests that idioms that do not exist in the higher-resource languages English and German are learned at the same pace as those with English or German counterparts. We find no evidence that the model relies on translation cues; idiomatic competence appears to stem from direct exposure rather than inference from English equivalents. Even in SmolLM$_\text{scratch}$, performance on Idioms$_\text{challenge}$ remains lower.

\paragraph{Error Analysis. }
We manually inspect the idioms that the best-performing model, AI Sweden LLaMA, does not prefer over the corrupted counterpart. A recurring pattern is that many of these idioms are infrequently used in contemporary Swedish. For instance, \textit{glad i hatten} (literally \say{happy in the hat}, meaning \say{drunk}) is widely recognized by Swedish speakers (according to an informal survey among friends and colleagues) but rarely used in active speech. Other idioms, such as \textit{få korgen} (literally \say{get the basket}, equivalent to the English \say{get the mitten}, meaning \say{to be turned down}), are unfamiliar to some speakers altogether, and considered old-fashioned by those who do know the expression. In such cases, it may be reasonable that the model assigns these expressions a low probability. We conclude that AI Sweden Llama learns Swedish idioms that are in active use today to a large extent. 

\subsubsection{Translationese}
\label{secsec:translationese}

The model's performance on the full Translationese set (Translationese$_\text{all}$) is extremely low --- well below chance --- indicating a strong bias toward more literal, translationese-like phrasing. However, since the Translationese examples tend to be shorter on average, this preference may reflect structural characteristics rather than genuine linguistic bias. For this reason, we focus our analysis on Translationese$_\text{filtered}$. 

\paragraph{Learning Curves.}

While Translationese$_\text{filtered}$ yields higher scores than Translationese$_\text{all}$, it still performs worst among all tasks. Notably, Translationese$_\text{filtered}$ (and even Translationese$_\text{all}$) reaches 95\% of its final performance earlier than any other dataset in training for the SmolLM models, as shown in Table~\ref{tab:95step}. Nonetheless, overall performance remains low, failing to surpass 70\%.

Interestingly, for the AI Sweden LLaMA model, 95\% of peak performance on the Translationese task is reached last among all tasks. We learn that \textit{if} (and only if) a model has sufficient capacity to learn to prefer idiomatic Swedish over Translationese, this process requires significantly longer adaptation time than learning to prefer grammatically correct Swedish. 

\paragraph{Error Analysis.}

Despite being the strongest model overall, AI Sweden LLaMA frequently assigns lower perplexity to clearly Translationese alternatives that would be dispreferred by native speakers. For example, in the sentence \say{I enjoy drawing}, the model favors the more literal translation \textit{Jag njuter av att rita}, where the verb \textit{njuta} is a direct equivalent of \say{enjoy}. However, the more idiomatic phrasing in Swedish is the alternative, \textit{Jag tycker om att rita}. Similarly, when translating \say{We only have money for noodles}, the model prefers the literal transfer of the expression, \textit{Vi har bara pengar till nudlar}, over the more natural expression \textit{Vi har bara råd med nudlar}.

In some cases, the preferred variants are not only unnatural but also incorrect. For instance, for \say{hidden costs}, the model selects \textit{gömda kostnader}, which is not used idiomatically in Swedish. The correct expression is \textit{dolda kostnader}, as the verb \textit{gömma} typically refers to the physical act of hiding an object, rather than abstract concealment. 

\subsection{Effect of Instruction Tuning (\ref{rq:ift})}
\label{sec:ift}

Across almost all tasks (with the exception of DaLaJ$_\text{punctuation}$, which we will discuss later in this section), we observe a decline in performance following instruction tuning, visible in Figures~\ref{fig:scratch-ift} and \ref{fig:cpt-ift}, and in Table~\ref{tab:final-accuracy}. The largest drop occurs on Translationese$_\text{filtered}$, indicating that the models' previously learned preference for idiomatic phrasings over Translationese is reversed again. For the SmolLM models, performance drops to the level of random guessing (50\%), suggesting that the ability to prefer idiomatic over literal translations is entirely lost during instruction tuning. 

For the larger AI Sweden LLaMA model, syntactic, morphological, and orthographic accuracy on ScaLA and DaLaJ remains largely unaffected. This suggests that core grammatical competence is robust to instruction tuning with translated data\footnote{At least for a language like Swedish, where machine translations produce mostly correct language, and likely even more standardized language than humans.}, provided the model has sufficient capacity. Lexical phenomena are somewhat more impacted, while the Idioms datasets and Translationese$_\text{filtered}$ degrade substantially.

This finding aligns with prior work showing that instruction tuning can harm general language modeling performance, as measured by perplexity \citep{jin2025demystifyinglanguagemodelforgetting, holmstrom-doostmohammadi-2023-making}. However, our results also identify which properties are most vulnerable to tuning with machine-translated data: fine-grained idiomatic competence is especially affected, while general linguistic abilities such as syntax and morphology remain relatively stable. 

In contrast to our results, \citet{waldis-etal-2024-holmes} even report \textit{improvements} in handling syntactic and morphological minimal pairs after instruction tuning. However, crucially, their work is on English, where instruction data is either natively written or generated by LLMs highly competent in English, and never machine-translated. Our findings highlights a key limitation of instruction tuning with machine-translated corpora. 

\paragraph{Punctuation} stands out in our results: As seen in Figure~\ref{fig:all-models-combined}, the accuracy on DaLaJ$_\text{punctuation}$ initially decreases during continued pre-training but \textit{increases} after tuning with translated data. This suggests that the punctuation standard encoded in the DaLaJ benchmark may align more closely with English norms than with common Swedish usage. Since this lies outside the primary scope of the paper, we leave an error analysis for future work. 

\paragraph{Implications. }
The learning (or rather, forgetting) curves in Figure~\ref{fig:all-models-combined} suggest that instruction tuning-data volume may not offer a simple trade-off between preserving linguistic competence and learning instruction-following behavior: The performance on the tasks probing for idiomatic language drops already \textit{early} in the instruction-tuning process. However, alternative regularizing hyperparameters, a lower learning rate, or replay of pre-training data in idiomatic Swedish may offer paths for preserving idiomatic skills, and are promising directions for future work. 

\subsection{Why is Idiomatic Language Sensitive?}
\label{sec:why}

As discussed in Section~\ref{secsec:idioms}, many of the idioms in our dataset are relatively rare in the pre-training data. We hypothesize that this is the result of idioms, and idiomatic language in general, being more prevalent in speech than in writing. It has long been established that speech differs fundamentally from text \citep{woolbert1922speaking}, with key differences driven by the expected level of formality, the degree of planning, and the number and familiarity of interlocutors \citep{Redeker01011984}. For instance, \citet{Allwood1998SomeFB} shows that spoken Swedish employs a smaller vocabulary and relies more heavily on pronouns, conjunctions, adverbs, and interjections, whereas written Swedish makes greater use of prepositions, nouns, and adjectives. 

This distinction is relevant not only for the Idioms dataset but also for the Translationese dataset, where certain idiomatic alternative translations occur more frequently in speech, while in many more formal forms of written text, one may prefer styles that differ less across languages. 

LLMs have been observed to favor a formal and impersonal style at least in English \citep{mitrović2023chatgpthumandetectexplain}. 
Despite the growing role of web-crawled data that includes many domains, encyclopedic and curated text sources remain a substantial part of pre-training. 
As a result, idiomatic language \textit{is} learned by our models, but more slowly. Instruction tuning then further amplifies this tendency: not only does it shift models toward a Translationese style due to the machine-translated data, but it also encourages more formal phrasing overall due to the structured, technical nature of the instructions and answers in the dataset. We speculate that this readjustment reduces the models’ probability to produce idiomatic alternatives.

\section{Conclusion}

In this work, we introduced two new datasets for measuring how idiomatic the language distribution of LLMs is for Swedish. The first dataset pairs conventionalized idioms with variants where a keyword in each expression is replaced by an alternative, often inspired by idioms in related higher-resourced languages. The second dataset contains sentence pairs contrasting Translationese Swedish with more idiomatic alternatives. 

Our first research question (\ref{rq:pretraining}) asked how idiomatic preferences are acquired during pre-training, compared to preferences for generally acceptable linguistic forms. Our results indicate that \say{baseline} linguistic skills such as syntactic and morphological preferences are learned early and converge quickly. Idiomatic language, by contrast, is harder to acquire and shows a slow, gradual improvement over the course of pre-training. 

We find no evidence that models infer conventionalized idioms from English equivalents: idioms that exist in both English and Swedish are learned at the same pace as Swedish-specific ones. Instead, English pretraining appears to provide a stronger foundation for memorizing idioms when they appear in the data: While models with English pretraining continue to improve over time, our model trained from scratch stops improving early and at a lower performance. 

Distinguishing Translationese from natural Swedish is the most challenging of our tasks. Even the strongest model tested often assigns a higher probability to literal, English-like phrasing even over clearly more idiomatic ones. 

Our second research question (\ref{rq:ift}) examined the influence of machine-translated instruction-tuning data on idiomatic preferences. We find that this setup substantially weakens models’ idiomatic competence already early in the tuning phase, while their performance on general linguistic acceptability probes remains comparatively stable. This suggests that idiomatic language is a particularly vulnerable ability. We therefore conclude that further work is needed to develop methods that preserve nuanced language skills in multilingual models lacking native instruction data.

\section*{Limitations}

\paragraph{Datasets.} The conclusions of this work are based on relatively small, manually curated datasets. As visible in the plots in Figure~\ref{fig:all-models-combined}, this leads to some variability in evaluation results. However, we observe that the \textit{trends} remain consistent across models, checkpoints and settings. 

Unfortunately, one of our datasets cannot be publicly released, as it is derived from copyrighted material. The source book is currently easily available for purchase, making it possible to reproduce the results, but given its niche nature, its availability may change over time. 

\paragraph{Coverage.} Our work focuses on Swedish, a mid-resource language closely related to English. Our findings that the acquisition of idiomatic language needs time and data is however particularly consequential for lower-resourced languages, where pre-training data is sparser and translations of instruction data will be worse. 
Those languages should therefore be the subject of future studies on idiomatic language acquisition. 

Finally, we restrict our analysis to small models due to the limited availability of large, open models with Swedish pretraining or continued pretraining and with training checkpoints available for download. Due to resource constraints on our side, we were \say{only} able to train 135M parameter models ourselves. Future work on larger models may yield insights into the acquisition of idiomatic language at different scales. 

\section*{Acknowledgments}

I thank TACL Action Editor Einat Minkov and the anonymous reviewers for their careful reading, constructive feedback, and insightful suggestions, which substantially improved this paper. I am grateful to student assistants Anja Jarochenko and Salome Kasendu for their help with annotating and revising the datasets in this study. Finally, I thank all colleagues who provided valuable feedback on edge-case idioms, in particular Karin Baardsen, Olle Torstensson, Oskar Holmström, Marco Kuhlmann, and Lukas Borggren. 
This research was supported by TrustLLM funded by Horizon Europe GA 101135671. The computations were enabled by the National Academic Infrastructure for Supercomputing in Sweden (NAISS), partially funded by the Swedish Research Council through grant agreement no. 2022-06725. 

\bibliographystyle{acl_natbib}
\bibliography{tacl2021}

\appendix
\section{Details about the Construction of the \textit{Idioms} Dataset}
\label{app:dataset_construction}

This appendix provides a detailed description of how the idiomatic minimal pairs were constructed, complementing the briefer description in Section~\ref{secsec:idioms_ds}. Its goal is to clarify how idiomatic keywords were identified ( §\ref{secsec:keyword_id}), how plausible but non-idiomatic alternatives were selected ( §\ref{secsec:const_alternatives}), and how these elements were embedded into full minimal-pair sentences ( §\ref{secsec:rewrite}). 

\subsection{Keyword Identification}
\label{secsec:keyword_id}

For each idiom, our annotators first identified a single idiomatic keyword that is essential to the conventionalized expression and that distinguishes it from plausible but non-idiomatic variants. This keyword is a content word that is required for the idiom.\footnote{The vast majority of the keywords are nouns, but even verbs, adjectives and adverbs occur.} Idioms that did not have a suitable content word were discarded from the dataset. 
Annotators then formulated a prompt sentence that explicitly conveys the idiom’s meaning or definition and is designed to elicit this keyword. These prompts are not intended to be an everyday usage of the idiom but rather to provide a clear semantic cue for the idiomatic expression. 

Consider this example from Section~\ref{secsec:idioms_ds}, for the idiom \textit{rik som ett troll}:\footnote{\url{https://sv.wiktionary.org/wiki/rik_som_ett_troll}} 
\begin{quote}
\textit{Någon som är väldigt rik är rik som ett troll.} \\
(\say{Someone who is very rich is rich like a troll.})
\end{quote}

Here, the idiomatic keyword is \textit{troll}, which is the central word of the conventionalized idiom \textit{rik som ett troll} (\say{rich like a troll}; \say{very rich}). 

\subsection{Construction of Plausible Alternatives}
\label{secsec:const_alternatives}

For each idiomatic keyword, we manually selected a distractor keyword that is semantically plausible in the same context but does not yield the idiomatic expression. Whenever possible, distractors were inspired by idioms in closely related high-resource languages (English or German). When no such counterpart existed, we selected a semantically compatible alternative from the same lexical category.

For the example above, the idiomatic keyword \textit{troll} was replaced by the 
distractor keyword \textit{kungabarn} (\say{prince} or \say{princess}), which is semantically equally plausible in context as both words convey the meaning \say{very rich}, but only \textit{rik som ett troll} forms a conventionalized idiom. 

Another example (also from Section~\ref{secsec:idioms_ds} is:\footnote{\url{https://sv.wiktionary.org/wiki/äpplen_och_päron})}
\begin{quote}
\textit{Om man jämför saker som inte är jämförbara, då jämför man äpplen och \ldots} \\
(\say{If one compares things that are not comparable, then one compares apples and \ldots}) \\
Keyword: \textit{Päron} (\say{Pears}) \\
Distractor: \textit{Apelsiner} (\say{Oranges})
\end{quote}
Here, the distractor \textit{apelsiner} (\say{oranges}) is chosen because there is a similar English idiom that goes \say{comparing apples and oranges}.

\subsection{Rewriting into Minimal Pair Sentences}
\label{secsec:rewrite}

To construct minimal pairs, we used the chatbot assistant Claude \citep{claude3} to rewrite the original prompt into two complete sentences that are identical except for the idiomatic keyword versus the distractor. The assistant was instructed to ensure that both variants were well-formed Swedish sentences and equal except for the keyword.
After this rewriting step, the idiomatic keyword (and its distractor) may appear in any position in the sentence, depending on what yields the most natural construction. This step results in minimal pairs suitable for full-sentence perplexity comparison. To give an example, the annotators, for the idiom \textit{ana ugglor i mossen} (\say{to sense owls in the bog}, \say{to sense that something is wrong}),\footnote{\url{https://sv.wiktionary.org/wiki/ana_ugglor_i_mossen}} wrote the following initial question and keyword: 
\begin{quote}
\textit{Vilka djur anar man i mossen när man misstänker att något är fel?} \\
(\say{Which animals does one sense in the bog when one suspects that something is wrong?})\\
Keyword: \textit{Ugglor} (\say{Owls}) \\ 
Distractor: \textit{Råttor} (\say{Rats})
\end{quote}
Then, the Claude model rewrote the example into the following minimal pair: 
\begin{quote}
\textit{När man misstänker att något är fel, anar man ugglor i mossen.} \\
(\say{When one suspects that something is wrong, one senses owls in the bog.”})

\textit{När man misstänker att något är fel, anar man råttor i mossen.} \\
(\say{When one suspects that something is wrong, one senses rats in the bog.})
\end{quote}

All generated sentence pairs were manually reviewed and, if necessary, edited by the annotators to ensure grammatical correctness, semantic equivalence between the two variants, and a clear preference for the idiomatic variant according to native-speaker judgments. 

\paragraph{Prompt} We used the following prompt to generate the minimal pairs: 
\begin{quote}
\textit{Construct two well-formed and coherent Swedish sentence alternatives from the following sentence by filling in the two completion words.  
The resulting sentences should be identical except for the completion word.}
\begin{itemize}
  \item \textit{Sentence:} [original sentence]
  \item \textit{Completion 1:} [keyword]
  \item \textit{Completion 2:} [distractor]
\end{itemize}
\end{quote}

\begin{table*}[h]
    \centering
    \adjustbox{max width=1\textwidth}{
    \begin{tabular}{l|l|ccc}\toprule
           Subset & Size of Subset & SmolLM$_\text{scratch}$ &SmolLM$_\text{CPT}$ & AI Sweden Llama\\\midrule
    All samples & 967 & 46.94 &  47.36 & 48.91   \\
    All filters, model tokenizer & 98 (SmolLM) / 134 (Llama) & 58.20 & 69.40 &  82.08   \\\midrule
    Manual only & 704 & 49.57  & 50.85 &   54.40 \\
    Length only, model tokenizer & 138 (SmolLM) / 185 (Llama) & 65.21 &  70.28 &  72.97 \\\midrule
    All filters, whitespace tokenizer & 285 & 56.84 & 62.10 & 61.40  \\
    Length only, whitespace tokenizer & 395 & 53.67 & 55.18 & 55.44  \\
    \bottomrule
    \end{tabular}
    }
    \caption{\textbf{Ablations on the Translationese subset filtering}. \textit{Manual only} includes examples where the idiomatic variant is clearly preferable according to general Swedish language norms, independent of additional context (see Section~\ref{secsec:translationese} for details). \textit{Length only} includes examples where the Translationese sample and the idiomatic alternative have the same length, determined either by whitespace tokenization or by the model tokenizer, as specified in the table. \textit{All filters} applies both criteria simultaneously.}
    \label{tab:translationese_ablations}
\end{table*}

\section{Ablations: Filtering of the \textit{Translationese} Subset}
\label{app:tokenizer}

The filtered Translationese dataset was constructed using two criteria: (i) sentence pairs must have the same length, and (ii) human annotators must agree that the Translationese sample is less idiomatic than its corrected counterpart. In this section, we ablate the two criteria to assess their individual impact.  

While human annotators excluded only 263 instances as not clearly Translationese, enforcing the strict length criterion eliminated 829 samples for SmolLM and 782 for Llama. For the best-performing AI Sweden Llama model, human filtering increased accuracy by $5.49$ points, whereas the length criterion increased accuracy by $24.06$ points (using the model tokenizer, as per our standard setup).  

\subsection{Effects of the Human Filtering}

As noted in the preceding paragraph, the human filtering leads to slightly higher scores, but does not do the heavy lifting in the subset. 
But although the length-matching constraint has a substantially larger practical influence, we hold on to the filtering because it impacts the quality of the dataset if some samples are not clearly identifiable as Translationese without more context. 

\subsection{Effects of the Tokenizer}

As noted in Section~\ref{subsubsub:translationese}, a limitation of basing the length restriction on the model tokenizer is that results cannot be directly compared across models with different tokenizers. While our focus in this paper is on intra-model comparisons (across model training checkpoints and before vs.\ after instruction tuning), this choice constrains the applicability of the dataset in comparisons of models with different tokenizers. In addition, the strict criterion substantially reduces the number of usable samples. To address this, we experiment with an alternative approach in which the length restriction is applied to whitespace-tokenized text instead.  

Table~\ref{tab:translationese_ablations} shows that relaxing the length criterion in this way yields a substantially larger dataset, with 285 samples retained compared to only 98 or 134 when using model-specific tokenizers. However, the bias towards the longer Translationese examples also remains stronger with the whitespace tokenizer: For example, for the AI Sweden Llama model, accuracy on the final subset is 82.08 when using the model tokenizer, but 61.40 when using the whitespace-based tokenizer. 

Finally, we would like to emphasize that both tokenizers for the models included in this study are English-centric. Tokenizers that are better adapted to Swedish, and thereby have a lower fertility for our dataset, will likely result in larger subsets. 

\end{document}